\documentclass{article}

  \PassOptionsToPackage{numbers, compress}{natbib}

\usepackage[final]{neurips_2024}

\usepackage[numbers]{natbib}
\usepackage[utf8]{inputenc} 
\usepackage[T1]{fontenc}    
\usepackage{hyperref}       
\usepackage{url}            
\usepackage{booktabs}       
\usepackage{amsfonts}       
\usepackage{nicefrac}       
\usepackage{microtype}      
\usepackage{xcolor}         

\usepackage{algorithm}
\usepackage{algorithmicx}
\usepackage[noend]{algpseudocode}
\usepackage{amssymb,amsmath,amsthm}
\usepackage[para,online,flushleft]{threeparttable}
\usepackage{mathtools}
\usepackage{multirow, makecell}
\usepackage{adjustbox}
\usepackage[normalem]{ulem}

\usepackage{caption}
\usepackage{subcaption}
\usepackage{wrapfig}
\usepackage{graphicx}
\usepackage{floatflt}

\title{Pruning-based Data Selection and Network Fusion for Efficient Deep Learning}

\author{%
  \textbf{Humaira Kousar\thanks{Equal Contribution. Correspondence to: Humaira Kousar <humairakousar32@kaist.ac.kr> \\ Department of Electrical Engineering, KAIST}} \\
  \\
  \And
\textbf{Hasnain Irshad Bhatti\footnotemark[1]}   \\\\
KAIST\\
  \And
 \textbf{Jaekyun Moon} \\
 \\
}

\makeatletter
\newcommand{\scriptsizecomments}[1]{\scriptsize\textcolor{black}{#1}} 
\algrenewcommand{\algorithmiccomment}[1]{\hfill\scriptsizecomments{#1}}
\makeatother

\begin{document}

\maketitle

\begin{abstract}

Efficient data selection is essential for improving the training efficiency of deep neural networks and reducing the associated annotation costs. However, traditional methods tend to be computationally expensive, limiting their scalability and real-world applicability. We introduce PruneFuse, a novel method that combines pruning and network fusion to enhance data selection and accelerate network training. In PruneFuse, the original dense network is pruned to generate a smaller surrogate model that efficiently selects the most informative samples from the dataset. Once this iterative data selection selects sufficient samples, the insights learned from the pruned model are seamlessly integrated with the dense model through network fusion, providing an optimized initialization that accelerates training. Extensive experimentation on various datasets demonstrates that PruneFuse significantly reduces computational costs for data selection, achieves better performance than baselines, and accelerates the overall training process.
\end{abstract}

\section{Introduction}
Deep learning models have achieved remarkable success across various domains, ranging from image recognition to natural language processing \citep{ren2015faster,long2015fully, he2016deep}. However, the performance of models heavily relies on the access of large amounts of labeled data for training \citep{sun2017revisiting}. In real-world applications, manually annotating massive datasets can be prohibitively expensive and time-consuming. Data selection techniques such as Active Learning (AL) \citep{gal2017deep} offer a promising solution to address this challenge by iteratively selecting the most informative samples from the unlabeled dataset for annotation. The goal of AL is to reduce labeling costs while maintaining or improving model performance. However, as data and modal complexity grow, traditional AL techniques that require iterative model training become computationally expensive, limiting scalability in resource-constrained settings. 

In this paper, we propose PruneFuse, a novel strategy for efficient data selection in active learning setting that overcomes the limitations of traditional approaches. Our approach is based on model pruning, which reduces the complexity of neural networks. By utilizing small pruned networks for data selection, we eliminate the need to train large models during the data selection phase, thus significantly reducing computational demands. Additionally after the data selection phase, we utilize the learning of these pruned networks to train the final model through a fusion process, which harnesses the insights from the trained networks to accelerate convergence and improve the generalization of the final model.

\textbf{Contributions.} Our key contribution is to introduce an efficient and rapid data selection technique that leverages pruned networks. By employing pruned networks as data selectors, PruneFuse ensures computationally efficient selection of informative samples leading to overall superior generalization. Furthermore, we propose a novel concept of fusing these pruned networks with the original untrained model, enhancing model initialization and accelerating convergence during training.

We demonstrate the broad applicability of PruneFuse across various network architectures, offering a flexible tool for efficient data selection in diverse deep learning settings. Extensive experimentation on CIFAR-10, CIFAR-100, and Tiny-ImageNet-200 shows that PruneFuse achieves superior performance to state-of-the-art active learning methods while significantly reducing computational costs.

\begin{figure}
    \vspace{-5mm}
    \centering
    \includegraphics[width=0.8\linewidth]{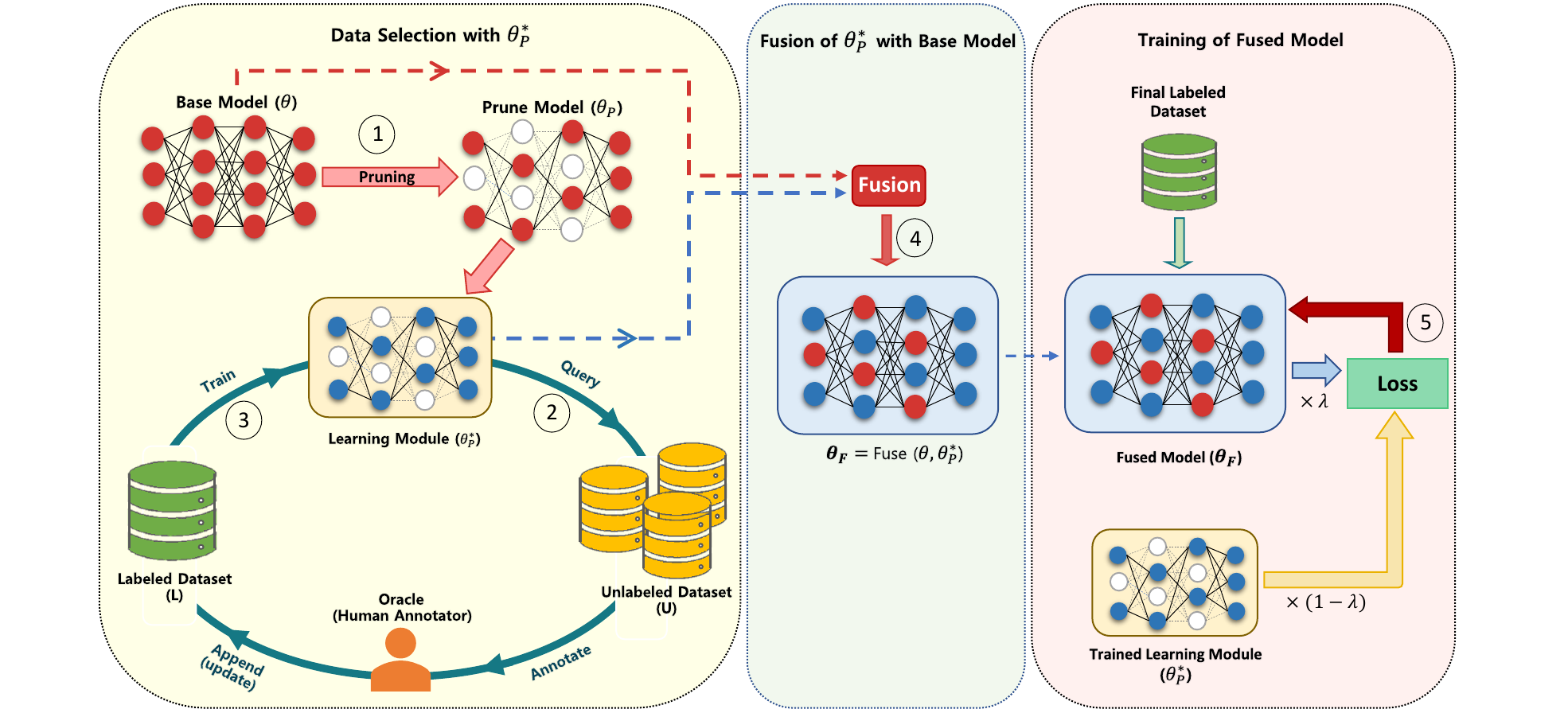}
    \vspace{-0.1cm}
    \caption{{\textbf{Overview of the PruneFuse Method}: (1) An untrained neural network is initially pruned to form a structured, pruned network $\theta_p$. (2) This pruned network $\theta_p$ queries the dataset to select prime candidates for annotation, similar to active learning techniques. (3) $\theta_p$ is then trained on these labeled samples to form the trained pruned network $\theta_p^*$. (4) The trained pruned network $\theta_p^*$ is fused with the base model $\theta$, resulting in a fused model. (5) The fused model is further trained on a selected subset of the data, incorporating knowledge distillation from $\theta_p^*$.}}
    \label{fig:our}
\vspace{-6.5mm}
\end{figure}

\section{Background and Related Works}

\textbf{Subset Selection Framework.}
Active Learning (AL) is widely utilized iterative approach tailored for situations with abundant unlabeled data. Given a classification task with \( C \) classes and a large pool of unlabeled samples \( U \), AL revolves around selectively querying the most informative samples from \( U \) for labeling. The process commences with an initial set of randomly sampled data \( s^0 \) from \( U \), which is subsequently labeled. In subsequent rounds, AL augments the labeled set \( L \) by adding newly identified informative samples. This cycle repeats until a predefined number of labeled samples $b$ are selected.

\textbf{Data Selection.} Approaches such as \citep{sener2017active, kirsch2019batchbald, gal2017deep} aim to select informative samples using techniques like diversity maximization and Bayesian uncertainty estimation. Parallelly, the domain of active learning has unveiled strategies, such as \citep{shen2017deep, sener2018active, kirsch2019batchbald, freytag2014selecting, kading2016active, sener2017active}, which prioritize samples that can maximize information gain, thereby enhancing model performance with minimal labeling effort. While these methods achieve efficient data selection, they still require training large models for the selection process, resulting in significant computational overhead. Other strategies such as \citep{killamsetty2021grad} optimize this selection process by matching the gradients of subset with training or validation set based on orthogonal matching algorithm and \citep{killamsetty2021glister} performs meta-learning based approach for online data selection. SubSelNet \citep{jain2024efficient} proposes to approximate a model that can be used to select the subset for various architectures without retraining the target model, hence reducing the overall overhead. However, it involves pre-training routine which is very costly and needed again for any change in data or model distribution. 
SVP \citep{coleman2019selection} introduces to use small proxy models for data selection but discards these proxies before training the target model. Additionally, structural discrepancies between the proxy and target models may result in sub-optimal data selections. Our approach also builds on this foundation of using small model (which in our case is a pruned model) but it enables direct integration with the target model through the fusion process. This ensures that the knowledge acquired during data selection is retained and actively contributes to the training of the original model. Also, the architectural coherence between the pruned and the target model provides a more seamless and effective mechanism for data selection, enhancing overall model performance and efficiency.

\textbf{Efficient Deep Learning.} Methods such as \citep{zoph2016neural, wan2020fbnetv2, han2015deep, dong2020hawq, jacob2018quantization, zhou2016dorefa, hinton2015distilling, yin2020dreaming} have been proposed to reduce model size and computational requirements.
Neural Network pruning has been extensively investigated as a technique to reduce the complexity of deep neural networks \citep{han2015deep}. Pruning strategies can be broadly divided into Unstructured Pruning \citep{park2016faster, dong2017learning, guo2016dynamic, park2020lookahead} and Structured Pruning \citep{li2016pruning, he2017channel, you2019gate, ding2019centripetal} based on the granularity and regularity of the pruning scheme. Unstructured pruning often yields a superior accuracy-size trade-off whereas structured pruning offers practical speedup and compression without necessitating specialized hardware \cite{liu2017learning}. While pruning literature suggests pruning after training \citep{renda2020comparing} or during training \citep{zhu2017prune, gale2019state},  recent research explore the viability of pruning at initialization \citep{lee2018snip, frankle2020pruning, tanaka2020pruning, frankle2020pruning, wang2020pruning}. In our work, we leverage the benefits of pruning at initialization to create a small representative model for efficient data selection.

\section{PruneFuse}
In this section, we delineate the PruneFuse methodology, illustrated in Fig. \ref{fig:our} (and Algorithm \ref{algo:algo_efficient_data_selection} provided in Appendix). The procedure begins with network pruning at initialization, offering a streamlined model for data selection. Upon attaining the desired data subset, the pruned model undergoes a fusion process with the original network, leveraging the structural coherence between them. The fused model is subsequently refined through knowledge distillation, enhancing its performance. We framed the problem as, let \( s_p \) be the subset selected using a pruned model \( \theta_p \) and \( s \) be the subset selected using the original model \( \theta \). We want to minimize:

\vspace{-3.5mm}
\begin{equation}
    \begin{aligned}
    \arg\min_{s_p} \left| E_{(x,y) \in s_p} [l(x, y; \theta, \theta_p)] - E_{(x,y) \in D} [l(x, y; \theta)]\right|  
    \end{aligned}
\end{equation}
Where \( E_{(x,y) \in s_p} [l(x, y; \theta, \theta_p)] \) is the expected loss on subset \( s_p \) (selected using \( \theta_p \)) when evaluated using the original model \( \theta \) and \( E_{(x,y) \in D} [l(x, y; \theta)] \) is the expected loss on full dataset $D$ when trained using the original model \( \theta \).  Furthermore,  the subset can be defined as $s_p = \left\{ (x_i, y_i) \in D \mid \text{score}(x_i, y_i; \theta_p) \geq \tau \right\}$ where
\(\text{score}(x_i, y_i; \theta_p)\) represents the score assigned to each sample selected using \( \theta_p \). The score function can be based on various strategies such as Least Confidence, Entropy, or Greedy k-centers. \(\tau\) defines the threshold used in the score-based selection methods (Least Confidence or Entropy) to determine the inclusion of a sample in \( s_p \).

The goal of the optimization problem is to select \( s_p \) such that when \( \theta \) is trained on it, the performance is as close as possible to training \( \theta \) on the full dataset \( D \). The key insight is that the subset \( s_p \) selected using the pruned model \( \theta_p \) is sufficiently representative and informative for training the original model \( \theta \). This is because \( \theta_p \) maintains a structure that is essentially identical to \( \theta \), although with some nodes pruned. As a result, there is a strong correlation between \( \theta \) and \( \theta_p \), ensuring that the selection made by \( \theta_p \) effectively minimizes the loss when \( \theta \) is trained on \( s_p \). By leveraging this surrogate \( \theta_p \), which is both computationally efficient and structurally coherent with \( \theta \), we can select most representative data out of $D$ to train \( \theta \).

\subsection{Pruning at Initialization}

Pruning at initialization \citep{wang2020pruning} shows potential in training time reduction, and enhanced model generalization. In our methodology, we employ structured pruning due to its benefits such as maintaining the architectural coherence of the network, enabling more predictable resource savings, and often leading to better-compressed models in practice.
Consider an untrained neural network, represented as \( \theta \). Let each layer \( \ell \) of this network have feature maps or channels denoted by \( c^{\ell} \), with \( \ell \in \{1, \ldots, L\} \). Channel pruning results in binary masks \( m^{\ell} \in \{0, 1\}^{d^{\ell}} \) for every layer, where \( d^{\ell} \) represents the total number of channels in layer \( \ell \). The pruned subnetwork, \( \theta_p \), retains channels described by \( c^{\ell} \odot m^{\ell} \), where \( \odot \) symbolizes the element-wise product. The sparsity $ p \in [0, 1]$ of the subnetwork illustrates the proportion of channels that are pruned: $p = 1 - \sum_{\ell} m^{\ell}/\sum_{\ell} d^{\ell}$.
To reduce the model complexity, we employ channel pruning procedure $prune(C, p)$. This prunes to a sparsity $p$ via two primary functions: i) score(C): This operation assigns scores $z^{\ell} \in \mathbb{R}^{d^{\ell}}$ to every channel in the network contingent on their magnitude (using the L2 norm). The channels $C$ are represented as $(c_1, \ldots, c_L)$. and ii) remove(Z, p): This process takes the magnitude scores $Z = (z_1, \ldots, z_L)$ and translates them into masks $m^{\ell}$ such that the cumulative sparsity of the network, in terms of channels, is $p$. We employ a one-shot channel pruning that scores all the channels simultaneously based on their magnitude and prunes the network from 0\% sparsity to $p\%$ sparsity in one cohesive step. Although previous works suggest re-initializing the network to ensure proper variance \citep{van2020single}. However, since the performance increment is marginal, we retain the weights of the pruned network before training.

\subsection{Data Selection via Pruned Model}

We begin by randomly selecting a small subset of data samples, denoted as $s^0$, from the unlabeled pool $U = \{x_i\}_{i\in[n]}$ where $[n] = \{1, . . . , n\}$. These samples are then annotated. The pruned model \( \theta_p \) is trained on this labeled subset \( s^0 \), resulting in the trained pruned model \( \theta_p^* \). With \( \theta_p^* \) as our tool, we venture into the larger unlabeled dataset \( U \) to identify samples that are prime candidates for annotation. Regardless of the scenario, our method employs three distinct criteria for data selection: Least Confidence (LC) \citep{Settles2012ActiveLearning}, Entropy \citep{shannon1948mathematical}, and Greedy k-centers \citep{sener2017active}. LC based selection gravitates towards samples where the pruned model exhibits the least confidence in its predictions. Thus, the uncertainty score for a given sample \( x_i \) is defined as $\text{score}(x_i; \theta_p)_{\text{LC}} = 1 - \max_{\hat{y}}P(\hat{y}|x_i; \theta_p^*)$. The entropy-based selection focuses on samples with high prediction entropy, computed as $\text{score}(x_i; \theta_p)_{\text{Entropy}} = - \sum_{\hat{y}} P(\hat{y}|x_i; \theta_p^*) \log P(\hat{y}|\mathbf{x}_i; \theta_p^*)$, highlighting uncertainty. Subsequently, we select the top-\( k \) samples exhibiting the highest uncertainty scores, proposing them as prime candidates for annotation. The Greedy k-centers aims to cherry-pick \( k \) centers from the dataset such that the maximum distance of any sample from its nearest center is minimized. The selection is mathematically represented as $x = \arg\max_{x \in U} \min_{c \in \text{centers}} d(x, c)$
where centers is the current set of chosen centers and $d(x,c)$ is the distance between point $x$ and center $c$. While various metrics can be employed to compute this distance, we opt for the Euclidean distance since it is widely used in this context.

\subsection{Training of Pruned Model}
Once we have selected the samples from \( U \), they are annotated to obtain their respective labels. These freshly labeled samples are assimilated into the labeled dataset \( L \). At the start of each training cycle, a fresh \( \theta_p \) is generated. Training from scratch in every iteration is vital to prevent the model from developing spurious correlations or overfitting to specific samples \citep{coleman2019selection}. This fresh start ensures that the model learns genuine patterns in the updated labeled dataset without carrying over potential biases from previous iterations. The training process adheres to a typical deep learning paradigm. Given the dataset \( L \) with samples \( (x_i, y_i) \), the aim is to minimize the loss function:
$\mathcal{L}(\theta_p, L) = \frac{1}{|L|} \sum_{i=1}^{|L|} \mathcal{L}_i(\theta_p, x_i, y_i)$, where \( \mathcal{L}_i \) denotes the individual loss for the sample \( x_i \). 
Training unfolds over multiple iterations (or epochs). In each iteration, the weights of \( \theta_p \) are updated using backpropagation with an optimization algorithm like stochastic gradient descent (SGD). 
This process is inherently iterative as in AL. After each round of training, new samples are chosen, annotated, and the model is reinitialized and retrained from scratch. This cycle persists until certain stopping criteria, e.g. labeling budget or desired performance, are met. With the incorporation of new labeled samples at every stage, \( \theta_p^* \) progressively refines its performance, becoming better suited for the subsequent data selection phase.

\subsection{Fusion with the Original Model}

\begin{wrapfigure}{r}{0.5\textwidth}
\vspace{-12mm}
\centering
\resizebox{6.5cm}{!}{
\begin{subfigure}[b]{0.30\linewidth}
    \centering
    \includegraphics[width=0.9\linewidth]{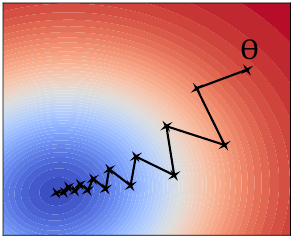}
    \caption{\scriptsizecomments{$\theta$ trajectory}}
    \label{subfig:theta_landscape}

    \includegraphics[width=0.9\linewidth]{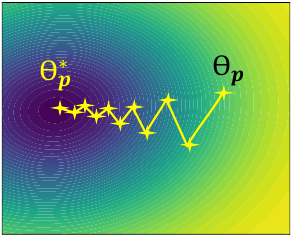}    
    \caption{\scriptsizecomments{$\theta_p$ trajectory}}
    \label{subfig:thetap_landscape}
  \end{subfigure}
  \begin{subfigure}[b]{0.70\linewidth}
    \centering
    \includegraphics[width=0.89\linewidth]{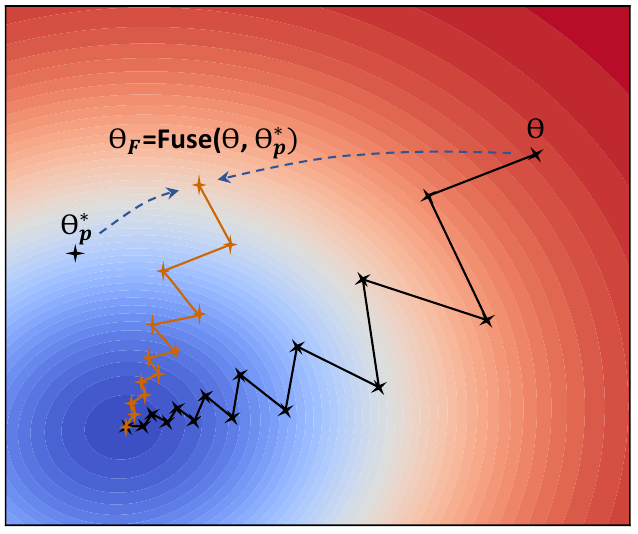}
    \caption{\scriptsizecomments{$\theta_F$ with a refined trajectory due to fusion}}
    \label{subfig:thetaf_landscape}
  \end{subfigure}}
  \caption{{\textbf{Evolution of training trajectories}. Pruning $\theta$ to $\theta_p$ tailors the loss landscape from \ref{subfig:theta_landscape} to \ref{subfig:thetap_landscape}, allowing $\theta_p$ to converge on an optimal configuration, denoted as $\theta^*_p$. This model, $\theta^*_p$, is later fused with the original $\theta$, which provides better initialization and offers superior trajectory for $\theta_F$ to follow, as depicted in \ref{subfig:thetaf_landscape}.}}
\vspace{-6mm}
\label{fusion_contour}
\end{wrapfigure}

After achieving the predetermined budget, the next phase is to integrate the insights from the trained pruned model \( \theta_p^* \) into the untrained original model \( \theta \). This step is crucial, as it amalgamates the learned knowledge from \( \theta_p^* \) with the expansive architecture of the original model, aiming to harness the best of both worlds.

\textbf{Rationale for Fusion.} Traditional pruning and fine-tuning methods often involve training a large model, pruning it down, and then fine-tuning the smaller model. While this is effective, it does not fully exploit the potential benefits of the larger, untrained model. The primary reason is that the pruning process might discard useful structures and connections within the original model that were not yet leveraged during initial training. By fusing the trained pruned model with the untrained original model, we aim to create a model that combines the learned knowledge by \( \theta_p^* \) with the broader, unexplored model \( \theta \).

\textbf{The Fusion Process.} Fusion is executed by transferring the weights from the trained pruned model's weight matrix \( \theta_p^* \) to the corresponding locations within the weight matrix of the untrained original model \( \theta \). This results in a new, fused weight matrix:
$
\theta_F = Fuse(\theta, \theta_p^*)
$.
Let's represent a model \( \theta \) as a sequence of layers, where each layer \( L \) consists of filters (for CNNs). We can denote the \(i^{th}\) filter of layer \(j\) in model \( \theta \) as \( F_{i,j}^{\theta} \). Given: \( \theta \) is the original untrained model and \( \theta_p^* \) is the trained pruned model. For a specific layer \( j \), \( \theta \) has a set of $n$ filters \( \{ F_{1,j}^{\theta}, F_{2,j}^{\theta}, ... F_{n,j}^{\theta} \} \) and \( \theta_p^* \) has a set of $m$ filters \( \{ F_{1,j}^{\theta_p^*}, F_{2,j}^{\theta_p^*}, ... F_{m,j}^{\theta_p^*} \} \) where \( m \leq n \) due to pruning. The fusion process for layer \( j \) can be mathematically represented as:
\[
F_{i,j}^{\theta_F} = 
\begin{cases} 
F_{i,j}^{\theta_p^*} & \text{if } F_{i,j}^{\theta_p^*} \text{ exists} \\
F_{i,j}^{\theta} & \text{otherwise} 
\end{cases}
\]
Where \( F_{i,j}^{\theta_F} \) is the \(i^{th}\) filter of layer \(j\) in the fused model \( \theta_F \). 

\textbf{Advantages of Retaining Unaltered Weights:} By copying weights from the trained pruned model \( \theta_p^* \) into their corresponding locations within the untrained original model \( \theta \), and leaving the remaining weights of \( \theta \) yet to be trained, we create a unique blend. The weights from \( \theta_p^* \) encapsulate the knowledge acquired during training, providing a foundation. Meanwhile, the rest of the untrained weights in \( \theta \) still have their initial values, offering an element of randomness. This duality fosters a richer exploration of the loss landscape during subsequent training. Fig. \ref{fusion_contour} illustrates the transformation in training trajectories resulting from the fusion process. The trained weights of \( \theta_p^* \) provides a better initialization, while the unaltered weights serve as gateways to unexplored regions in the loss landscape. This strategic combination in the fused model \( \theta_F\) enables the discovery of potentially superior solutions that neither the pruned nor the original model might have discovered on their own.

\subsection{Refinement via Knowledge Distillation}

After the fusion process, our resultant model, \( \theta_F \), embodies a synthesis of insights from both the trained pruned model \( \theta_p^* \) and the original model \( \theta \). Although PruneFuse outperforms baseline AL  (results are provided in Appendix), we further optimize and enhance $\theta_F$ using Knowledge Distillation (KD). KD enables $\theta_F$ to learn from $\theta_p^*$ (the teacher model), enriching its training.
During the fine-tuning phase, we use two losses: i) Cross-Entropy Loss, which quantifies the divergence between the predictions of \( \theta_F \) and the actual labels in dataset \( L \), and ii) Distillation Loss, which measures the difference in the softened logits of \( \theta_F \) and \( \theta_p^* \). These softened logits are derived by tempering logits of \( \theta_p^* \), which in our case is the teacher model, with a temperature parameter before applying the softmax function. The composite loss is formulated as a weighted average of both losses. The iterative enhancement of \( \theta_F \) is governed by:
$
\theta_F^{(t+1)} = \theta_F^{(t)} - \alpha \nabla_{\theta_F^{(t)}} \left( \lambda \mathcal{L}_{\text{Cross Entropy}}(\theta_F^{(t)}, L) + (1-\lambda) \mathcal{L}_{\text{Distillation}}(\theta_F^{(t)}, \theta_p^*) \right)
$.
Here \( \alpha \) represents the learning rate, while \( \lambda \) functions as a coefficient to balance the contributions of the two losses. By incorporating KD in the fine-tuning phase, we aim to ensure that the fused model \( \theta_F \) not only retains the trained weights of pruned model but also reinforce this knowledge iteratively, optimizing the performance of \( \theta_F \) in subsequent tasks.

\begin{table*}[!t]
     \centering
  \resizebox{14cm}{!}{
\begin{tabular}{c||c|ccccc|ccccc||c|ccccc}
    \toprule
       \multirow{4}{*}{\textbf{Method}}& \multirow{4}{*}{\textbf{Params}} & \multicolumn{5}{c|}{\textbf{CIFAR-10}} & \multicolumn{5}{c||}{\textbf{CIFAR-100}} & \multirow{4}{*}{\textbf{Params}}&\multicolumn{5}{c}{\textbf{Tiny-ImageNet-200}}\\ \cmidrule{3-8}\cmidrule{9-12}\cmidrule{14-18}
       
          &   & \multicolumn{5}{c|}{\textbf{Label Budget ($b$)}} &  \multicolumn{5}{c||}{\textbf{Label Budget ($b$)}}&&\multicolumn{5}{c}{\textbf{Label Budget ($b$)}} \\ \cmidrule{3-8}\cmidrule{9-12}\cmidrule{14-18}
        
       &(Million)   & \textbf{10\%}  & \textbf{20\%}  & \textbf{30\%}  & \textbf{40\%}  & \textbf{50\%}& \textbf{10\%}  & \textbf{20\%}  & \textbf{30\%}  & \textbf{40\%}  & \textbf{50\%} &(Million)   & \textbf{10\%}  & \textbf{20\%}  & \textbf{30\%}  & \textbf{40\%}  & \textbf{50\%}\\ \midrule

        Baseline \small{$(AL)$}  
        &0.85  &  80.53 & 87.74 & 90.85 & 92.24 & 93.00&  
        35.99 & 52.99 & 59.29 & 63.68 & 66.72& 
        25.56& 14.86 & 33.62 & 43.96 & 49.86	& 54.65\\

        \midrule
     
        PruneFuse \small{$(p=0.5)$}
        &0.21 & \textbf{80.92} & \textbf{88.35} & \textbf{91.44} & \textbf{92.77} & \textbf{93.65} & \textbf{40.26} & \textbf{53.90} & \textbf{60.80} & \textbf{64.98} & \textbf{67.87} & 6.10 &   \textbf{18.71} &	\textbf{39.70} &	\textbf{47.41} &	\textbf{51.84} & \textbf{55.89}\\

        \midrule
        
         PruneFuse \small{$(p=0.6)$}
          &0.13 & \textbf{80.58} & 87.79 & 90.94 & \textbf{92.58} & \textbf{93.08} & \textbf{37.82}& 52.65 & \textbf{60.08} & \textbf{63.7} & \textbf{66.89} & 3.92 & \textbf{19.25} & \textbf{38.84} & \textbf{47.02} & \textbf{52.09} & \textbf{55.29}\\

        \midrule

         PruneFuse \small{$(p=0.7)$} 
       & 0.07 & 80.19 & \textbf{87.88} & 90.70 & \textbf{92.44} & \textbf{93.40} & \textbf{36.76} & 52.15 & \textbf{59.33} & 63.65 & \textbf{66.84}& 2.23 & \textbf{18.32} & \textbf{39.24} & \textbf{46.45} & \textbf{52.02} & \textbf{55.63}\\

        \midrule

         PruneFuse \small{$(p=0.8)$}     
          &0.03 & 80.11 & 87.58 & 90.50 & \textbf{92.42} & \textbf{93.32} & \textbf{36.49} & 50.98 & 58.53 & 62.87 & 65.85 & 1.02 &  \textbf{18.34}	&\textbf{37.86}&	\textbf{47.15}	&\textbf{51.77}	&\textbf{55.18}\\

        \midrule

    \end{tabular}
}
     \vspace{-2mm}
 \caption{{\textbf{Performance Comparison} of Baseline and PruneFuse on CIFAR-10, CIFAR-100 and Tiny ImageNet-200. This table summarizes the test accuracy of final models (original in case of AL and Fused in PruneFuse) for various pruning ratios ($p$) and labeling budgets($b$). Least Confidence is used as a metric for subset selection and different architectures (ResNet-56 for CIFAR-10 and CIFAR-100 while ResNet-50 for Tiny-ImageNet) are utilized.}}
 \vskip -2.5mm
  \label{tab:main_Acc_R56}
\end{table*}

\begin{figure}[!t]
\begin{center}
\begin{tabular}{cccc}
\includegraphics[width=0.2255\textwidth]{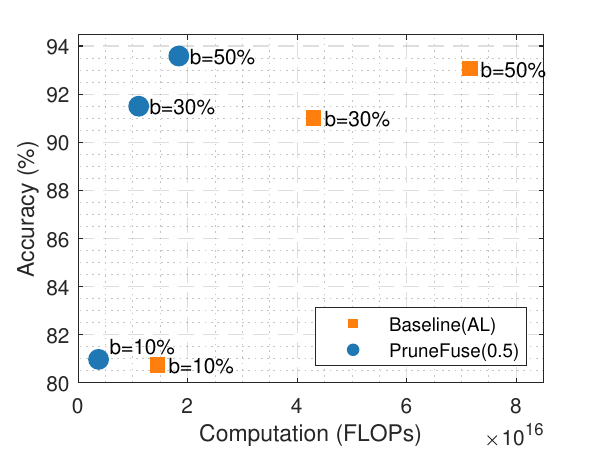} &
\includegraphics[width=0.2255\textwidth]{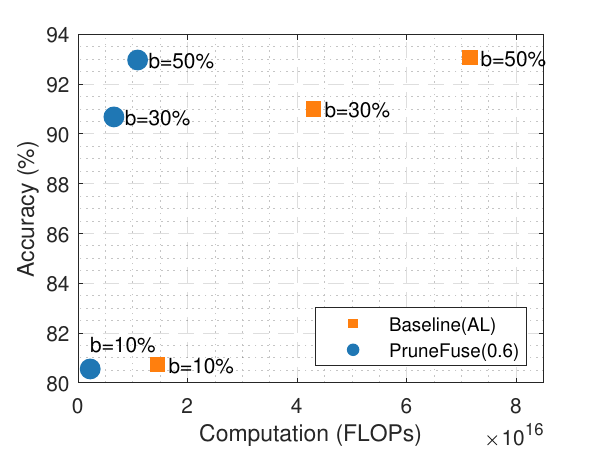} &
\includegraphics[width=0.2255\textwidth]{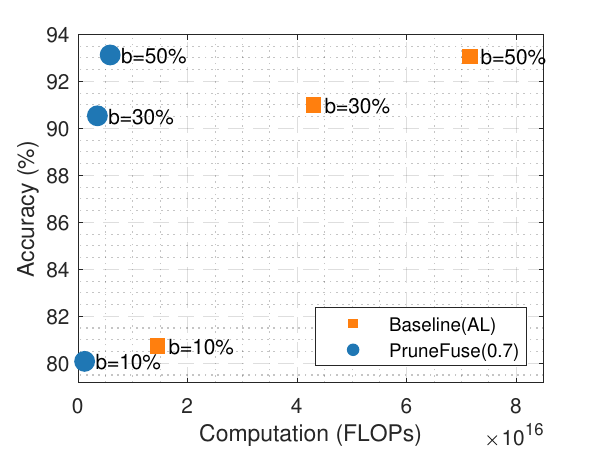}&
\includegraphics[width=0.2255\textwidth]{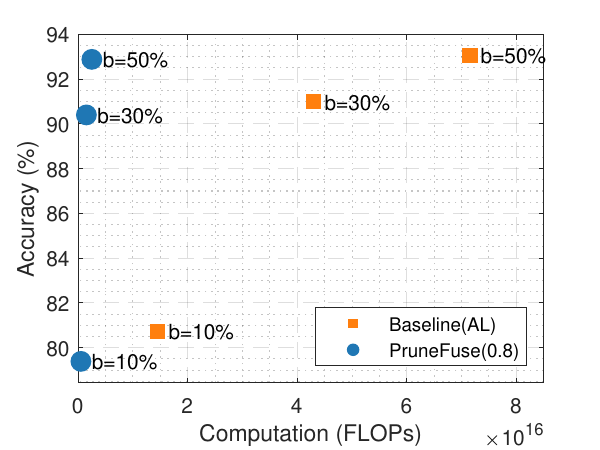}\\

\scriptsizecomments{(a) } & \scriptsizecomments{(b) }& \scriptsizecomments{(c) }& \scriptsizecomments{(d)} 

\end{tabular}
\vspace{-2mm}
\caption{{ \textbf{Computation Comparison of PruneFuse and Baseline (Active Learning):} This figure illustrates the total number of FLOPs utilized by PruneFuse, compared to the baseline Active Learning method, for selecting subsets with specific labeling budgets $b=10\%, 30\%, 50\%$. The experiments are conducted on the CIFAR-10 dataset using the ResNet-56 architecture. Subfigures (a), (b), (c), and (d) correspond to different pruning ratios (0.5, 0.6, 0.7, and 0.8, respectively)}.}
\label{fig:computation}
\end{center}
\end{figure}

\section{Experiments}

\vspace{-2mm}
\textbf{Experimental Setup. }
The effectiveness of our approach is assessed on three image classification datasets; CIFAR-10 \citep{krizhevsky2009learning}, CIFAR-100 \citep{krizhevsky2009learning}, and TinyImageNet-200 \citep{le2015tiny}.
We used ResNet-50, ResNet-56 and ResNet-164 architecture in our experiments. We pruned these architectures using the Torch-Prunnig library \citep{fang2023depgraph} for different pruning ratios $p=$ $0.5, 0.6, 0.7,$ and $0.8$ to get the pruned architectures. We trained the model for 181 epochs using the mini-batch of 128 for CIFAR-10 and CIFAR-100 and 100 epochs using the mini-batch of 256 for TinyImageNet-200. For all the experiments SGD is used as an optimizer. We took AL as a baseline for the proposed technique and initially, we started by randomly selecting 2\% of the data. For the first round, we added 8\% from the unlabeled set, then 10\% in each subsequent round, until reaching the label budget, $b$. After each round, we retrained the models from scratch, as described in the methodology.
All experiments are carried out independently 3 times and then the average is reported. 


\subsection{Results and Discussions}

\textbf{Main Experiments.} Table \ref{tab:main_Acc_R56} summarizes the generalization performance of baseline and different variants of PruneFuse on different datasets (detailed results on different architectures and data selection metrics are provided in Appendix). All variants of PruneFuse achieve higher accuracy compared to the baseline, demonstrating the effectiveness of superior data selection performance and fusion. Fig. \ref{fig:computation} (a), (b), (c), and (d) illustrates the computational complexity of the baseline and PruneFuse variants in terms of Floating Point Operations (FLOPs) for different labeling budgets. The FLOPs are computed for the whole training duration of the pruned network and the selection process. Different variants of PruneFuse $p=0.5, 0.6, 0.7,$ and $0.8$ provide the flexibility that the user can choose the variant of PruneFuse depending on their computation resources e.g. PruneFuse ($p=0.8$) requires very low computation resources compared to others while achieving good accuracy performance.



\begin{table}
\centering
 \vspace{-3.5mm}
 \resizebox{10.5cm}{!}{
\begin{tabular}{c|c|c|c||ccccc}
\toprule
\multirow{2}{*}{\textbf{Method}} & \multirow{2}{*}{\textbf{Model}}&\multirow{2}{*}{\textbf{Architecture}} & \multirow{2}{*}{\textbf{No. of Parameters}}& \multicolumn{5}{c}{\textbf{Label Budget ($b$)}} \\ \cmidrule{5-9}

& &&\small{(Million)} &\textbf{10\%} & \textbf{20\%} & \textbf{30\%} & \textbf{40\%} & \textbf{50\%} \\ \midrule

\multirow{2}{*}{\textbf{SVP}} & Data Selector & ResNet-20  & 0.26  & 81.07 & 86.51 & 89.77 & 91.08 & 91.61 \\\cmidrule{2-9}
                           &  Target &  ResNet-56 &     0.85    & 80.76 & 87.31 & 90.77 & 92.59 & 92.95 \\ \midrule
                           
\multirow{2}{*}{\textbf{PruneFuse}} &Data Selector & ResNet-56 ($p=0.5$)) & \textbf{0.21} & 78.62 & 84.92 & 88.17 & 89.93 & 90.31 \\\cmidrule{2-9}
                            & Target &ResNet-56&   0.85   & \textbf{82.68} & \textbf{88.97} & \textbf{91.63} & \textbf{93.24} & \textbf{93.69}\\
\bottomrule
\end{tabular}}
\vspace{1.5mm}
 \caption{ Performance Comparison of SVP and PruneFuse across various labeling budgets $b$ for efficient training of Target Model (ResNet-56).}
  \label{tab:Svp vs pruneFuse}

\vskip -4mm
\end{table}

\begin{table}[t]
    \centering

\resizebox{7.25cm}{!}{
\begin{tabular}{cc||ccccc}
\toprule
\multirow{2}{*}{\textbf{Method}}  & \multirow{2}{*}{\textbf{Selection Metric}}& \multicolumn{5}{c}{\textbf{Label Budget ($b$)}} \\ \cmidrule{3-7}

 & &\textbf{10\%} & \textbf{20\%} & \textbf{30\%} & \textbf{40\%} & \textbf{50\%} \\ \midrule

       \multirowcell{4}{Baseline \\ \small{$AL$}} 

        & Least Conf & 35.99 & 52.99 & 59.29 & 63.68 & 66.72 \\  
        ~ & Entropy & 37.57 & 52.64 & 58.87 & 63.97 & 66.78 \\ 
        ~ & Random & 37.06 & 51.62 & 58.77 & 62.05 & 64.63 \\ 
        ~ & Greedy k & 38.28 & 52.43 & 58.96 & 63.56 & 66.3 \\
        \midrule
        \multirowcell{4}{PruneFuse \\ \small{$p=0.5$}\\(without KD)} 

        & Least Conf & \textbf{39.27} & \textbf{54.25} & \textbf{60.6} & \textbf{64.17} & \textbf{67.49} \\
        & Entropy & \textbf{37.43} & \textbf{52.57} & \textbf{60.57} & \textbf{64.44} & \textbf{67.31} \\
        & Random & \textbf{40.07} & \textbf{52.83} & \textbf{59.93} & \textbf{63.06} & \textbf{65.41} \\
        & Greedy k & \textbf{39.25} & \textbf{52.43} & \textbf{59.94} & \textbf{63.94} & \textbf{66.56} \\

                \midrule
        \multirowcell{4}{PruneFuse \\ \small{$p=0.5$}\\(with KD)} 
        & Least Conf & \textbf{40.26} & \textbf{53.90} & \textbf{60.80} & \textbf{64.98} & \textbf{67.87} \\ 
        & Entropy & \textbf{38.59} & \textbf{54.01} & \textbf{60.52} & \textbf{64.83} & \textbf{67.67} \\ 
        & Random & \textbf{39.43} & \textbf{54.60} & \textbf{60.13} & \textbf{63.91} & \textbf{66.02} \\ 
        & Greedy k  & \textbf{39.83} & \textbf{54.35} & \textbf{60.40} & \textbf{64.22} & \textbf{66.89} \\ 
    
\bottomrule
\end{tabular}}
\vspace{1.5mm}
\caption{ Ablation Study of Knowledge Distillation on PruneFuse for CIFAR-100 datasets on Resnet-56}
\label{tab:KD_Abla}
\vspace{-4mm}
\end{table}

  
  
  
  

\textbf{Comparison with Selection-via-Proxy.} Table \ref{tab:Svp vs pruneFuse} delineates a comparison of PruneFuse and the SVP \cite{coleman2019selection}, performance metrics show that PruneFuse consistently outperforms SVP across all labeling budgets for the efficient training of a Target Model (ResNet-56). 
 SVP employs a ResNet-20 as its data selector, with a model size of 0.26 M. In contrast, PruneFuse uses a 50\% pruned ResNet-56, reducing its data selector size to 0.21 M. Notably, while the data selector of PruneFuse achieves a lower accuracy of 90.31\% at $b=50\%$ compared to SVP's 91.61\%, the target model utilizing PruneFuse-selected data attains a superior accuracy of 93.69\%, relative to 92.95\% for the SVP-selected data. This disparity underscores the distinct operational focus of the data selectors: PruneFuse's selector is optimized for enhancing the target model's performance, rather than its own.




\textbf{Ablation Studies.}
Table \ref{tab:KD_Abla} demonstrates the effect of Knowledge Distillation (KD) on the PruneFuse technique relative to the baseline method across various data selection matrices and label budgets on  CIFAR-100 datasets, using ResNet-56 architecture. The results indicate that PruneFuse consistently outperforms the baseline method, both with and without incorporating KD from a trained pruned model. This superior performance is attributed to the innovative fusion strategy inherent to PruneFuse. The proposed approach gives the fused model an optimized starting point, enhancing its ability to learn more efficiently and generalize better. The impact of this strategy is evident across different label budgets and architectures, demonstrating its effectiveness and robustness.

\begin{wrapfigure} {r}{0.5\textwidth}
\vspace{-7mm}
\begin{center}
\resizebox{7.25cm}{!}{
\begin{tabular}{ccc}
 \includegraphics[width=0.25\textwidth]{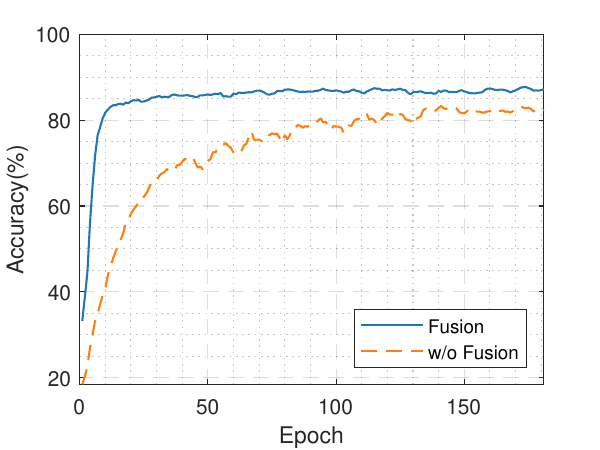} &
\includegraphics[width=0.25\textwidth]{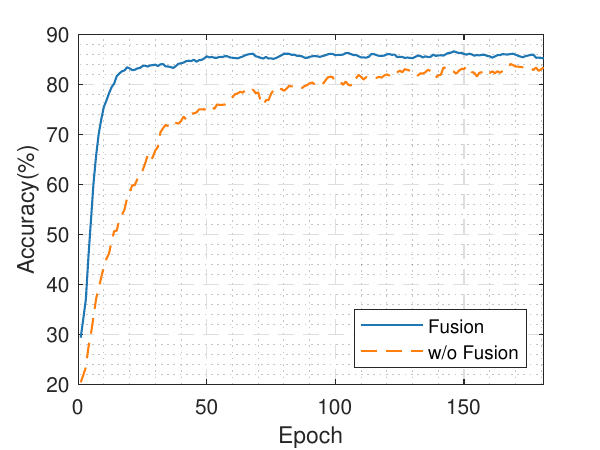}\\

\scriptsizecomments{(a) $p=0.5$, $b=30\%$ } & \scriptsizecomments{(b) $p=0.6$, $b=30\%$}

\end{tabular}}
\caption{{\textbf{Impact of Model Fusion on PruneFuse Performance:} This figure compares the accuracy over epochs between fused and non-fused training approaches within the PruneFuse framework, both utilizing subset (with labeling budget $b$) selected by the pruned model. Experiments are conducted using the ResNet-56 on the CIFAR-10. Subfigures (a) and (b) correspond to pruning ratios $p=0.5$ and $0.6,$ respectively.}}
\label{fig:Effect_of_fusion}
\end{center}
\vspace{-4mm}
\end{wrapfigure}

Fig. \ref{fig:Effect_of_fusion} demonstrates the effect of fusion across various pruning ratios, the models trained with fusion in-place perform better than those trained without fusion, achieving higher accuracy levels at an accelerated pace. The rapid convergence is most notable in initial training phases, where fusion model benefits from the initialization provided by the integration of weights from a trained pruned model $\theta_p^*$ with an untrained model $\theta$. The strategic retention of untrained weights introduces a beneficial stochastic component to the training process, enhancing the model's ability to explore new regions of the parameter space. This dual capability of exploiting prior knowledge and exploring new configurations enables the proposed technique to consistently outperform, making it particularly beneficial in scenarios with sparse label data.

\vspace{-3mm}
\section{Conclusion}
\vspace{-3mm}
We introduce PruneFuse, a novel approach combining pruning and network fusion to optimize data selection in deep learning. PruneFuse leverages a small pruned model for data selection, which then seamlessly fuses with the original model, providing fast and better generalization while significantly reducing computational costs. Extensive evaluations on CIFAR-10, CIFAR-100, and Tiny-ImageNet-200 show that PruneFuse outperforms existing baselines, establishing its efficiency and efficacy.
PruneFuse offers a scalable, practical, and flexible solution to enhance the training efficiency of neural networks, particularly in resource-constrained settings.


\section{Acknowledgments}
This work was supported by the National Research Foundation of Korea (NRF) grant funded by the Korea government
(MSIT) (No. RS-2024-00340966), and by Center for Applied Research in Artificial Intelligence (CARAI)
grant funded by DAPA and ADD (UD230017TD).

\bibliography{neurips_2024}

\newpage
\appendix
\section{Appendix}

\subsection{Performance Comparison with different Datasets, Selection Metrics, and Architectures}
To comprehensively evaluate the effectiveness of PruneFuse, we conducted additional experiments comparing its performance with baseline utilizing other data selection metrics such as Least Confidence, Entropy, and Greedy k-centers. Results are shown in Tables \ref{supp:tab:main_Acc_R56_Selection} and \ref{supp:tab:Acc_Resnet164} for various architectures and labeling budgets. In all cases, our results demonstrate that PruneFuse mostly outperforms the baseline using these traditional metrics across various datasets and model architectures, highlighting the robustness of PruneFuse in selecting the most informative samples efficiently. 

\begin{table*}[h]
     \centering
  \resizebox{14cm}{!}{
  \subfloat[CIFAR-10.]{\begin{tabular}{cc||ccccc}
    \toprule
       
          \multirow{2}{*}{\textbf{Method}} & \multirow{2}{*}{\textbf{Selection Metric}} & \multicolumn{5}{c}{\textbf{Label Budget ($b$)}}  \\ \cmidrule{3-7}
        
        &  & \textbf{10\%}  & \textbf{20\%}  & \textbf{30\%}  & \textbf{40\%}  & \textbf{50\%} \\ \midrule

        \multirowcell{4}{Baseline \\ \small{$AL$}}  
        & Least Conf &  80.53 & 87.74 & 90.85 & 92.24 & 93.00\\
        & Entropy & 80.14 &	87.63 &	90.80 &	92.51 &	92.98 \\
        & Random & 78.55 &	85.26 &	88.13 &	89.81 &	91.20 \\
        & Greedy k & 79.63 &	86.46 &	90.09 &	91.9 &	92.80 \\
        \midrule
        \multirowcell{4}{PruneFuse \\ \small{$p=0.5$}} 
        & Least Conf & 80.92 & 88.35 & 91.44 & 92.77 & 93.65 \\ 
        & Entropy & 81.08 & 88.74 & 91.33 & 92.78 & 93.48 \\ 
        & Random & 80.43 & 86.28 & 88.75 & 90.36 & 91.42 \\ 
        & Greedy k  & 79.85 & 86.96 & 90.20 & 91.82 & 92.89 \\ 
        \midrule
        \multirowcell{4}{PruneFuse \\ \small{$p=0.6$}} 
        & Least Conf & 80.58 & 87.79 & 90.94 & 92.58 & 93.08 \\
        & Entropy & 80.96 & 87.89 & 91.22 & 92.56 & 93.19 \\
        ~ & Random & 79.19 & 85.65 & 88.27 & 90.13 & 91.01 \\
        ~ & Greedy k & 79.54 & 86.16 & 89.5 & 91.35 & 92.39 \\
        \midrule
        \multirowcell{4}{PruneFuse \\ \small{$p=0.7$}} 
        & Least Conf & 80.19 & 87.88 & 90.70 & 92.44 & 93.40 \\
        & Entropy & 79.73 & 87.85 & 90.94 & 92.41 & 93.39 \\
        
        ~ & Random & 78.76 & 85.5 & 88.31 & 89.94 & 90.87 \\
        ~ & Greedy k & 78.93 & 85.85 & 88.96 & 90.93 & 92.23 \\
        \midrule
        \multirowcell{4}{PruneFuse \\ \small{$p=0.8$}}     
          & Least Conf & 80.11 & 87.58 & 90.50 & 92.42 & 93.32 \\
        ~ & Entropy & 79.83 & 87.5 & 90.52 & 92.24 & 93.15 \\
        ~ & Random & 78.77 & 85.64 & 88.45 & 89.88 & 91.21 \\
        ~ & Greedy k & 78.23 & 85.59 & 88.60 & 90.11 & 91.31 \\
        \midrule

    \end{tabular}}

\quad

\subfloat[CIFAR-100.]{\begin{tabular}{cc||ccccc}
    \toprule
       
          \multirow{2}{*}{\textbf{Method}} & \multirow{2}{*}{\textbf{Selection Metric}} & \multicolumn{5}{c}{\textbf{Label Budget ($b$)}}  \\ \cmidrule{3-7}
        
        &  & \textbf{10\%}  & \textbf{20\%}  & \textbf{30\%}  & \textbf{40\%}  & \textbf{50\%} \\ \midrule

        \multirowcell{4}{Baseline \\ \small{$AL$}} 
        & Least Conf &  35.99 & 52.99 & 59.29 & 63.68 & 66.72 \\
        & Entropy &37.57 & 52.64 & 58.87 & 63.97 & 66.78 \\
        & Random & 37.06 & 51.62 & 58.77 & 62.05 & 64.63 \\
        & Greedy k & 38.28 & 52.43 & 58.96 & 63.56 & 66.30 \\

        \midrule
        \multirowcell{4}{PruneFuse \\ \small{$p=0.5$}} 
        & Least Conf & 40.26 & 53.90 & 60.80 & 64.98 & 67.87 \\ 
        & Entropy & 38.59 & 54.01 & 60.52 & 64.83 & 67.67 \\ 
        & Random & 39.43 & 54.60 & 60.13 & 63.91 & 66.02 \\ 
        & Greedy k  & 39.83 & 54.35 & 60.40 & 64.22 & 66.89 \\ 

        \midrule
        \multirowcell{4}{PruneFuse \\ \small{$p=0.6$}} 
         & Least Conf & 37.82 & 52.65 & 60.08 & 63.7 & 66.89 \\
        & Entropy & 38.01 & 51.91 & 59.18 & 63.53 & 66.88 \\
        & Random & 38.27 & 52.85 & 58.68 & 62.28 & 65.2 \\
        & Greedy k & 38.44 & 52.85 & 59.36 & 63.36 & 66.12 \\
        
        \midrule

        \multirowcell{4}{PruneFuse \\ \small{$p=0.7$}} 
        & Least Conf & 36.76 & 52.15 & 59.33 & 63.65 & 66.84 \\
        ~ & Entropy & 36.95 & 50.64 & 58.45 & 62.27 & 65.88 \\
        ~ & Random & 37.3 & 51.66 & 58.79 & 62.67 & 65.08 \\
        ~ & Greedy k & 38.88 & 52.02 & 58.66 & 61.39 & 65.28 \\
        \midrule
        \multirowcell{4}{PruneFuse \\ \small{$p=0.8$}}     
          & Least Conf & 36.49 & 50.98 & 58.53 & 62.87 & 65.85 \\
        ~ & Entropy & 36.02 & 51.23 & 57.44 & 62.65 & 65.76 \\
        ~ & Random & 37.37 & 52.06 & 58.19 & 62.19 & 64.77 \\
        ~ & Greedy k & 37.04 & 49.84 & 56.13 & 60.24 & 62.92 \\
        \midrule

    \end{tabular}}}
 \caption{\textbf{Performance Comparison} of Baseline and PruneFuse on CIFAR-10 and CIFAR-100 with ResNet-56 architecture. This table summarizes the test accuracy of final models (original in case of AL and Fused in PruneFuse) for various pruning ratios ($p$), labeling budgets ($b$), and data selection metrics.}
 \vskip -4mm
  \label{supp:tab:main_Acc_R56_Selection}
\end{table*}

\begin{table*}[h]
     \centering
  \vspace{0.6cm}
  \resizebox{14cm}{!}{
  \subfloat[CIFAR-10.]{\begin{tabular}{cc||ccccc}
    \toprule
       
          \multirow{2}{*}{\textbf{Method}} & \multirow{2}{*}{\textbf{Selection Metric}} & \multicolumn{5}{c}{\textbf{Label Budget ($b$)}}  \\ \cmidrule{3-7}
        
        &  & \textbf{10\%}  & \textbf{20\%}  & \textbf{30\%}  & \textbf{40\%}  & \textbf{50\%} \\ \midrule

        \multirowcell{4}{Baseline \\ \small{$AL$}} 
        & Least Conf. &  81.15 & 89.4 & 92.72 & 94.09 & 94.63\\
        & Entropy &  80.99 & 89.54 & 92.45 & 94.06 & 94.49 \\
        & Random & 80.27 & 87.00 & 89.94 & 91.57 & 92.78 \\
        & Greedy k & 80.02 & 88.33 & 91.76 & 93.39 & 94.40\\
        \midrule
        \multirowcell{4}{PruneFuse \\ \small{$p=0.5$}} 

        & Least Conf. & 83.03 & 90.30 & 93.00 & 94.41 & 94.63 \\ 
        ~ & Entropy & 82.64 & 89.88 & 93.08 & 94.32 & 94.90 \\ 
        ~ & Random & 81.52 & 87.84 & 90.14 & 91.94 & 92.81 \\ 
        ~ & Greedy k & 81.70 & 88.75 & 91.92 & 93.64 & 94.22 \\ 
        \midrule
        \multirowcell{4}{PruneFuse \\ \small{$p=0.6$}} 

        & Least Conf. & 82.86 & 90.22 & 93.05 & 94.27 & 94.66 \\ 
        ~ & Entropy & 82.23 & 90.18 & 92.91 & 94.28 & 94.66 \\ 
        ~ & Random & 81.14 & 87.51 & 90.05 & 91.82 & 92.43 \\ 
        ~ & Greedy k & 81.11 & 88.41 & 91.66 & 92.94 & 94.17 \\ 
        \midrule
        \multirowcell{4}{PruneFuse \\ \small{$p=0.7$}} 
        & Least Conf. & 82.76 & 89.89 & 92.83 & 94.10 & 94.69 \\ 
        ~ & Entropy & 82.59 & 89.81 & 92.77 & 94.20 & 94.74 \\ 
        ~ & Random & 80.88 & 87.54 & 90.09 & 91.57 & 92.64 \\ 
        ~ & Greedy k & 81.68 & 88.36 & 91.64 & 93.02 & 93.97 \\ 
        \midrule
        \multirowcell{4}{PruneFuse \\ \small{$p=0.8$}} 
          & Least Conf. & 82.66 & 89.78 & 92.64 & 94.08 & 94.69 \\ 
        ~ & Entropy & 82.01 & 89.77 & 92.65 & 94.02 & 94.60 \\ 
        ~ & Random & 80.73 & 87.43 & 90.08 & 91.40 & 92.53 \\ 
        ~ & Greedy k & 79.66 & 87.56 & 90.79 & 92.30 & 93.17 \\ 
        \midrule

    \end{tabular}}
     \vspace{4mm}

\quad

\subfloat[CIFAR-100.]{\begin{tabular}{cc||ccccc}
    \toprule
       
          \multirow{2}{*}{\textbf{Method}} & \multirow{2}{*}{\textbf{Selection Metric}} & \multicolumn{5}{c}{\textbf{Label Budget ($b$)}}  \\ \cmidrule{3-7}
        
        &  & \textbf{10\%}  & \textbf{20\%}  & \textbf{30\%}  & \textbf{40\%}  & \textbf{50\%} \\ \midrule

       \multirowcell{4}{Baseline \\ \small{$AL$}}

        & Least Conf & 38.41 & 51.39 & 65.53 & 70.07 & 73.05 \\ 
        ~ & Entropy & 36.65 & 57.58 & 64.98 & 69.99 & 72.90 \\ 
        ~ & Random & 39.31 & 57.53 & 63.84 & 67.75 & 70.79 \\ 
        ~ & Greedy k & 39.76 & 57.40 & 65.20 & 69.25 & 72.91 \\ 
        \midrule
        \multirowcell{4}{PruneFuse \\ \small{$p=0.5$}} 

        & Least Conf & 42.88 & 59.31 & 66.95 & 71.45 & 74.32 \\ 
        ~ & Entropy & 42.99 & 59.32 & 66.83 & 71.18 & 74.43 \\ 
        ~ & Random & 43.72 & 58.58 & 64.93 & 68.75 & 71.63 \\ 
        ~ & Greedy k & 43.61 & 58.38 & 66.04 & 69.83 & 73.10 \\ 
        \midrule
        \multirowcell{4}{PruneFuse \\ \small{$p=0.6$}} 

        & Least Conf & 41.86 & 58.97 & 66.61 & 70.59 & 73.6 \\ 
        ~ & Entropy & 42.43 & 58.74 & 65.97 & 70.90 & 73.70 \\ 
        ~ & Random & 42.53 & 58.33 & 65.00 & 68.55 & 71.46 \\ 
        ~ & Greedy k & 42.71 & 58.41 & 65.43 & 69.57 & 72.49 \\ 
        \midrule
        \multirowcell{4}{PruneFuse \\ \small{$p=0.7$}} 

         & Least Conf & 42.00 & 57.08 & 66.41 & 70.68 & 73.63 \\ 
        ~ & Entropy & 41.01 & 57.45 & 65.99 & 70.07 & 73.45 \\ 
        ~ & Random & 42.76 & 57.31 & 64.12 & 68.07 & 70.88 \\ 
        ~ & Greedy k & 42.42 & 57.58 & 65.18 & 68.55 & 71.89 \\ 

        \midrule
        \multirowcell{4}{PruneFuse \\ \small{$p=0.8$}} 

            & Least Conf & 41.19 & 57.98 & 65.22 & 70.38 & 73.17 \\ 
        ~ & Entropy & 39.78 & 57.3 & 65.19 & 69.40 & 72.82 \\ 
        ~ & Random & 42.08 & 57.23 & 64.05 & 67.85 & 70.62 \\ 
        ~ & Greedy k & 42.20 & 57.42 & 64.53 & 68.01 & 71.29 \\ 
        
        \midrule

    \end{tabular}}}
 \caption{\textbf{Performance Comparison} of Baseline and PruneFuse on CIFAR-10 and CIFAR-100 with ResNet-164 architecture. This table summarizes the test accuracy of final models (original in case of AL and Fused in PruneFuse) for various pruning ratios ($p$), labeling budgets ($b$), and data selection metrics.}
  \label{supp:tab:Acc_Resnet164}
\end{table*}

\newpage
\subsection{Comparison with SVP}

\begin{wraptable}{r}{0.5\textwidth}
\centering
 \vspace{-5mm}
 \resizebox{6.89cm}{!}{
\begin{tabular}{c|c|c|c||ccccc}
\toprule
\multirow{2}{*}{\textbf{Techniques}} & \multirow{2}{*}{\textbf{Model}}&\multirow{2}{*}{\textbf{Architecture}} & \multirow{2}{*}{\textbf{No. of Parameters}}& \multicolumn{5}{c}{\textbf{Label Budget ($b$)}} \\ \cmidrule{5-9}

& &&\small{(Million)} &\textbf{10\%} & \textbf{20\%} & \textbf{30\%} & \textbf{40\%} & \textbf{50\%} \\ \midrule
\multirow{2}{*}{\textbf{SVP}} &  Data Selector& ResNet-8 & 0.074  & 77.85 & 83.35 & 85.43 &	86.83 & 86.90 \\\cmidrule{2-9}
         & Target &  ResNet-20 &     0.26   & 80.18 & 86.34	& 89.22	&90.75	&91.88 \\ \midrule
\multirow{2}{*}{\textbf{PruneFuse}} &  Data Selector& ResNet-20 ($p=0.5$) & \textbf{0.066} & 76.58	&83.41	&85.83	&87.07	&88.06 \\\cmidrule{2-9}
         & Target&ResNet-20&  0.26  & \textbf{80.25}&	\textbf{87.57}	&\textbf{90.20}&	\textbf{91.70}&	\textbf{92.29}\\
\bottomrule
\end{tabular}}
\caption{Comparison of SVP and PruneFuse on Small Models.}
\label{supp:table:SvpvspruneFuseonsmallmodels}
\vspace{-3mm}
\end{wraptable}

Table \ref{supp:table:SvpvspruneFuseonsmallmodels} demonstrates the performance comparison of PruneFuse and SVP for small model architecture ResNet-20 on CIFAR-10. SVP achieves 91.88\% performance accuracy by utilizing the data selector having 0.074 M parameters whereas PruneFuse outperforms SVP by achieving 92.29\%  accuracy with a data selector of 0.066 M parameters.

\begin{wrapfigure}{r}{0.5\textwidth}
\centering
\vspace{-1.65cm}
\resizebox{6.89cm}{!}{
\begin{tabular}{cc}
\includegraphics[width=0.26\textwidth,  trim={3.5cm 8cm 3.5cm 8cm}, clip]{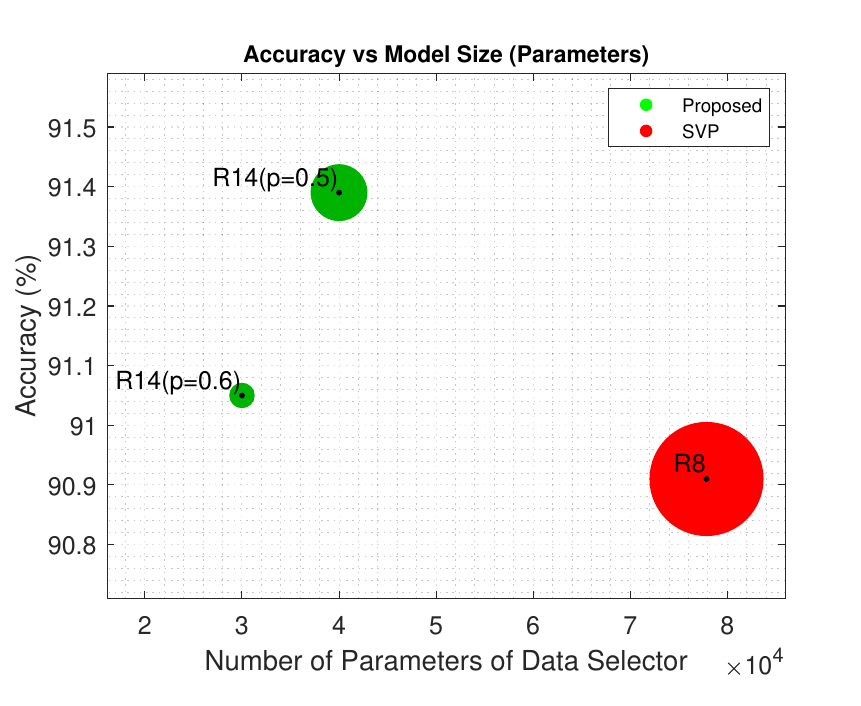} &

\includegraphics[width=0.26\textwidth, trim={3.5cm 8cm 3.5cm 8cm}, clip]{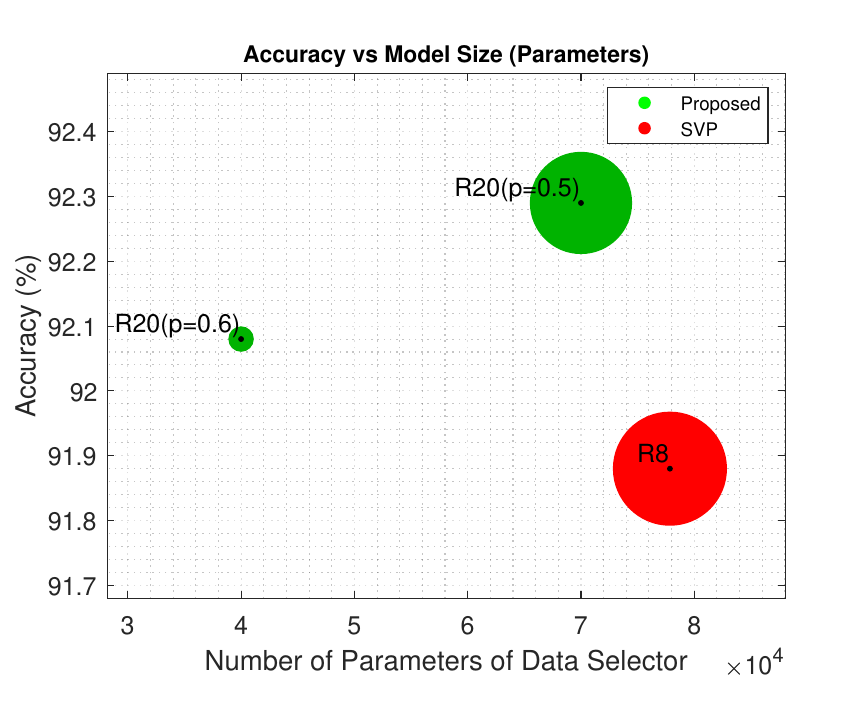}\\
\vspace{-1mm}
\scriptsizecomments{(a) Target Model = ResNet-14} & \scriptsizecomments{(b) Target Model = ResNet-20} \\
\end{tabular}}
\vspace{-1mm}
\caption{\textbf{Comparison of PruneFuse with SVP.} \scriptsize{Scatter plot shows final accuracy on target model against the model size for different ResNet models on CIFAR-10 dataset with labeling budget $b =$ 50\%. (a) shows for the target network ResNet-14, ResNet-14 (with $p=0.5$ and $p=0.6$) and ResNet-8 models are used as data selectors for PruneFuse and SVP, respectively. While in (b), PruneFuse utilizes ResNet20 (i.e. $p=0.5$ and $p=0.6$) and SVP utilizes ResNet-8 models for data selection when the target model is ResNet-20. }}
\label{fig:scaterplot_svp_Vs_proposed}

 \vskip -5mm
\end{wrapfigure}

Fig. \ref{fig:scaterplot_svp_Vs_proposed}(a) and (b) show that target models when trained with the data selectors of the PruneFuse achieve significantly higher accuracy while using significantly less number of parameters compared to SVP. These results indicate that the PruneFuse does not require an additional architecture for designing the data selector; it solely needs the target model. In contrast, SVP necessitates both the target model (ResNet-14) and a smaller model (ResNet-8) that functions as a data selector.

\subsection{Ablation Study of Fusion}
\begin{figure}[!h]
\begin{center}
\begin{tabular}{ccc}

\includegraphics[width=0.3\textwidth]{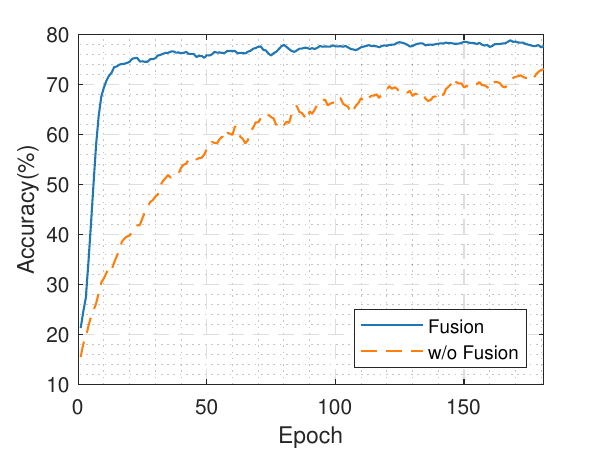} &
\includegraphics[width=0.3\textwidth]{init_ablation_prune3_2.pdf} &
\includegraphics[width=0.3\textwidth]{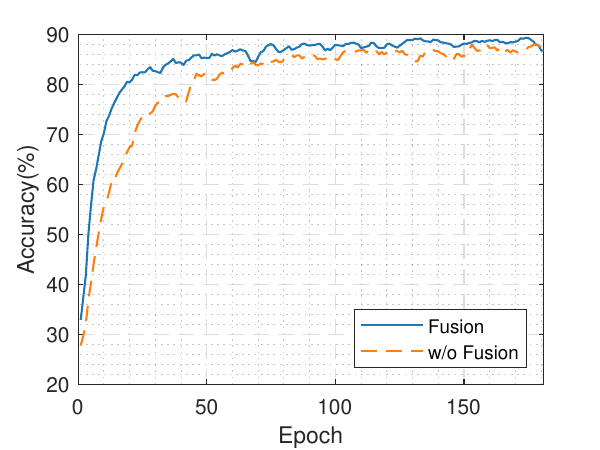}\\

\footnotesize{(a) $p=0.5$, $b=10\%$} & \footnotesize{(b) $p=0.5$, $b=30\%$}& \footnotesize{(c) $p=0.5$, $b=50\%$}

\end{tabular}

\caption{ \textbf{Ablation Study of Fusion on PruneFuse} ($p=0.5$). Experiments are performed on ResNet-56 architecture with CIFAR-10.}
\label{supp:fusion0.5}
\end{center}

\vskip -3mm
\end{figure}

The fusion process is a critical component of the PruneFuse methodology, designed to integrate the knowledge gained by the pruned model into the original network. Our experiments reveal that models trained with the fusion process exhibit significantly better performance and faster convergence compared to those trained without fusion. By initializing the original model with the weights from the trained pruned model, the fused model benefits from an optimized starting point, which enhances its learning efficiency and generalization capability. Fig. \ref{supp:fusion0.5} illustrates the training trajectories and accuracy improvements when fusion takes places, demonstrating the tangible benefits of this initialization. These results underscore the importance of the fusion step in maximizing the overall performance of the PruneFuse framework.

\subsection{Ablation Study of Knowledge Distillation in PruneFuse}
Table \ref{tab:KD_Ablation} demonstrates the effect of Knowledge Distillation on the PruneFuse technique relative to the baseline Active Learning (AL) method across various experimental configurations and label budgets on CIFAR-10 and CIFAR-100 datasets, using ResNet-56 architecture. The results indicate that PruneFuse consistently outperforms the baseline method, both with and without incorporating Knowledge Distillation (KD) from a trained pruned model. This superior performance is attributed to the innovative fusion strategy inherent to PruneFuse, where the original model is initialized using weights from a previously trained pruned model. The proposed approach gives the fused model an optimized starting point, enhancing its ability to learn more efficiently and generalize better. The impact of this strategy is evident across different label budgets and architectures, demonstrating its effectiveness and robustness. 

\begin{table*}[h]
     \centering
  \vspace{0.6cm}
  \resizebox{14cm}{!}{
  \subfloat[CIFAR-10.]{\begin{tabular}{cc||ccccc}
\toprule
        \multirow{2}{*}{\textbf{Method}}  & \multirow{2}{*}{\textbf{Selection Metric}}& \multicolumn{5}{c}{\textbf{Label Budget ($b$)}} \\ \cmidrule{3-7}
    
         & &\textbf{10\%} & \textbf{20\%} & \textbf{30\%} & \textbf{40\%} & \textbf{50\%} \\ \midrule
    
       \multirowcell{4}{Baseline \\ \small{$AL$}} 

        & Least Conf & 80.53 & 87.74 & 90.85 & 92.24 & 93.00 \\   
        ~ & Entropy & 80.14 & 87.63 & 90.80 & 92.51 & 92.98 \\ 
        ~ & Random & 78.55 & 85.26 & 88.13 & 89.81 & 91.20 \\ 
        ~ & Greedy k &  79.63 & 86.46 & 90.09 & 91.90 & 92.80 \\ 
        
        \midrule
        \multirowcell{4}{PruneFuse \\ \small{$p=0.5$}\\(without KD)} 

        & Least Conf & \textbf{81.08} & \textbf{88.71} & \textbf{91.24} & \textbf{92.68} & \textbf{93.46} \\
        & Entropy & \textbf{80.80} & \textbf{88.08} & \textbf{90.98} & \textbf{92.74} & \textbf{93.43} \\
        & Random & \textbf{80.11} & \textbf{85.78} & \textbf{88.81} & \textbf{90.20} & 91.10 \\

        & Greedy k & \textbf{80.07} & \textbf{86.70} & 89.93 & 91.72 & 92.67 \\

                \midrule
        \multirowcell{4}{PruneFuse \\ \small{$p=0.5$}\\(with KD)} 
        & Least Conf & \textbf{80.92} & \textbf{88.35} & \textbf{91.44} & \textbf{92.77} & \textbf{93.65} \\ 
        & Entropy & \textbf{81.08} & \textbf{88.74} & \textbf{91.33} & \textbf{92.78} & \textbf{93.48} \\ 
       & Random & \textbf{80.43} & \textbf{86.28} & \textbf{88.75} & \textbf{90.36} & \textbf{91.42} \\
        & Greedy k & \textbf{79.85} & \textbf{86.96} & \textbf{90.20} & 91.82 & \textbf{92.89} \\
        \bottomrule
\end{tabular}}
    \vspace{4mm}
\quad

\subfloat[CIFAR-100.]{\begin{tabular}{cc||ccccc}
\toprule
\multirow{2}{*}{\textbf{Method}}  & \multirow{2}{*}{\textbf{Selection Metric}}& \multicolumn{5}{c}{\textbf{Label Budget ($b$)}} \\ \cmidrule{3-7}

 & &\textbf{10\%} & \textbf{20\%} & \textbf{30\%} & \textbf{40\%} & \textbf{50\%} \\ \midrule

       \multirowcell{4}{Baseline \\ \small{$AL$}} 

        & Least Conf & 35.99 & 52.99 & 59.29 & 63.68 & 66.72 \\  
        ~ & Entropy & 37.57 & 52.64 & 58.87 & 63.97 & 66.78 \\ 
        ~ & Random & 37.06 & 51.62 & 58.77 & 62.05 & 64.63 \\ 
        ~ & Greedy k & 38.28 & 52.43 & 58.96 & 63.56 & 66.3 \\
        \midrule
        \multirowcell{4}{PruneFuse \\ \small{$p=0.5$}\\(without KD)} 

        & Least Conf & \textbf{39.27} & \textbf{54.25} & \textbf{60.6} & \textbf{64.17} & \textbf{67.49} \\
        & Entropy & \textbf{37.43} & \textbf{52.57} & \textbf{60.57} & \textbf{64.44} & \textbf{67.31} \\
        & Random & \textbf{40.07} & \textbf{52.83} & \textbf{59.93} & \textbf{63.06} & \textbf{65.41} \\
        & Greedy k & \textbf{39.25} & \textbf{52.43} & \textbf{59.94} & \textbf{63.94} & \textbf{66.56} \\

                \midrule
        \multirowcell{4}{PruneFuse \\ \small{$p=0.5$}\\(with KD)} 
        & Least Conf & \textbf{40.26} & \textbf{53.90} & \textbf{60.80} & \textbf{64.98} & \textbf{67.87} \\ 
        & Entropy & \textbf{38.59} & \textbf{54.01} & \textbf{60.52} & \textbf{64.83} & \textbf{67.67} \\ 
        & Random & \textbf{39.43} & \textbf{54.60} & \textbf{60.13} & \textbf{63.91} & \textbf{66.02} \\ 
        & Greedy k  & \textbf{39.83} & \textbf{54.35} & \textbf{60.40} & \textbf{64.22} & \textbf{66.89} \\ 
    
\bottomrule
\end{tabular}}}

\caption{ Ablation Study of Knowledge Distillation on PruneFuse presented in (a), and (b) for different datasets on Resnet-56}
\label{tab:KD_Ablation}
\end{table*}

\subsection{Ablation of Different Selection Metrics}
The impact of different selection metrics is presented in Table \ref{tab:selectionMatric_Ablation} demonstrating clear differences in performance across both the Baseline and PruneFuse methods on CIFAR-100 using ResNet-164 architecture. It is evident that across both the baseline and PruneFuse methods, the Least Confidence metric surfaces as particularly effective in optimizing label utilization and model performance. The results further reinforce that regardless of the label budget, from 10\% to 50\%, PruneFuse demonstrates a consistent superiority in performance with different data selection metrics (Least Confidence, Entropy, Random, and Greedy k-centres) compared to  Baseline. 

\begin{table}[h]
\centering
 \resizebox{7cm}{!}{
\begin{tabular}{cc||ccccc}
\toprule
\multirow{2}{*}{\textbf{Method}}  & \multirow{2}{*}{\textbf{Selection Metric}}& \multicolumn{5}{c}{\textbf{Label Budget ($b$)}} \\ \cmidrule{3-7}

 & &\textbf{10\%} & \textbf{20\%} & \textbf{30\%} & \textbf{40\%} & \textbf{50\%} \\ \midrule

       \multirowcell{4}{Baseline \\ \small{$AL$}}

        & Least Conf & 38.41 & 51.39 & 65.53 & 70.07 & 73.05 \\ 
        ~ & Entropy & 36.65 & 57.58 & 64.98 & 69.99 & 72.90 \\ 
        ~ & Random & 39.31 & 57.53 & 63.84 & 67.75 & 70.79 \\ 
        ~ & Greedy k & 39.76 & 57.40 & 65.20 & 69.25 & 72.91 \\ 
        \midrule
        \multirowcell{4}{PruneFuse \\ \small{$p=0.5$}} 

        & Least Conf & \textbf{42.88} & \textbf{59.31} & \textbf{66.95} & \textbf{71.45} & \textbf{74.32} \\ 
        ~ & Entropy & \textbf{42.99} & \textbf{59.32} & \textbf{66.83} & \textbf{71.18} & \textbf{74.43} \\ 
        ~ & Random & \textbf{43.72} & \textbf{58.58} & \textbf{64.93} & \textbf{68.75} & \textbf{71.63} \\ 
        ~ & Greedy k & \textbf{43.61} & \textbf{58.38} & \textbf{66.04} & \textbf{69.83} & \textbf{73.10} \\ 
                           
\bottomrule
\end{tabular}}
\vspace{1mm}
 \caption{Effect of Different Data Selection Metrics on CIFAR-100 using ResNet-164 architecture.}
  \label{tab:selectionMatric_Ablation}
  \vskip -5mm
\end{table}

\section{Algorithmic Details}
In this section, we provide a detailed explanation of the PruneFuse algorithm given in Algorithm \ref{algo:algo_efficient_data_selection}. The PruneFuse methodology begins with structured pruning an untrained neural network, $\theta$, to create a smaller model, $\theta_p$. This pruning step reduces complexity while retaining the network's essential structure, allowing $\theta_p$ to efficiently select informative samples from the unlabeled dataset, $U$. The algorithm proceeds as follows. First, the original model $\theta$ is randomly initialized and pruned to obtain $\theta_p$. The pruned model $\theta_p$ is then trained on an initial labeled dataset $s^0$ to produce $\theta_p^*$. This training equips $\theta_p$ with preliminary knowledge for data selection. The labeled dataset $L$ is initially set to $s^0$. A data selection loop runs until the labeled dataset $L$ reaches the maximum budget $b$. In each iteration, $\theta_p$ is retrained on $L$ to keep the model updated with new samples. Uncertainty scores for all samples in $U$ are computed using the trained $\theta_p^*$ on the available labelel subset, and the top-$k$ samples with the highest scores are selected as $D_k$. These samples are labeled and added to $L$. Once the budget $b$ is met, the final trained pruned model $\theta_p^*$ is fused with the original model $\theta$ to create the fused model $\theta_F$. This fusion transfers the weights from $\theta_p^*$ to $\theta$, ensuring the pruned model's knowledge is retained. Finally, $\theta_F$ is trained on $L$ using knowledge distillation from $\theta_p^*$, refining the model's performance by leveraging the pruned model's learned insights. In summary, PruneFuse strategically adapts pruning in data selection problem and to enhance both data selection efficiency and model performance.

\begin{algorithm}[h]

\caption{Efficient Data Selection using Pruned Networks}
\label{algo:algo_efficient_data_selection}
\textbf{Input}: Unlabeled dataset $U$, Initial labeled dataset $s^0$, labeled dataset $L$, original model $\theta$, prune model $\theta_p$, fuse model $\theta_F$, maximum budget $b$, pruning ratio $p$.\\

\begin{algorithmic}[1]
\State Randomly initialize $\theta $
\State $\theta_p \gets$ Prune$(\theta,p)$ \hfill// Structure pruning

\State $\theta_p^*  \gets$ Train  $\theta_p $ on $s^0$
\State $L \gets s^0$

\vspace{0.5em}
\While{$|L| \leq b$}

\State Compute ${\text{score}}(\mathbf{x}; \theta_p^*)$ for all $x \in U$ \hfill //Compute uncertainty scores for samples in $U$ using $\theta_p^*$



\State $D_k = top_k[D_j \in U]_{j \in [k]}$ \hfill //Select top-$k$ samples with highest uncertainty scores
\State Query labels $y_k$ for selected samples $D_k$
\State Add $(D_k,y_k)$ to $L$
\State Train  $\theta_p^* $ on $L$

\EndWhile
\vspace{0.5em}
\State $\theta_F \gets Fuse(\theta , \theta_p^*$) 
\State $\theta_F^* \gets $ Train $\theta_F$ on $L$
\vspace{0.5em}
\State\Return {$L$, $\theta_F^*$}
\end{algorithmic}
\end{algorithm}


\clearpage

\end{document}